\pdfoutput=1
\documentclass[letterpaper,10pt,conference]{ieeeconf}

\IEEEoverridecommandlockouts                              

\usepackage{graphics} 
\usepackage{graphicx}
\usepackage{mathptmx} 
\usepackage{amsmath} 
\usepackage{amssymb}  
\usepackage{tabularx}
\usepackage{adjustbox}
\usepackage{subcaption}
\usepackage[font=small, justification=justified]{caption}
\usepackage{verbatim}
\usepackage[
top    = 0.75in,
bottom = 0.75in,
left   = 0.75in,
right  = 0.75in]{geometry}

\newcommand{\norm}[1]{\left\lVert#1\right\rVert}

\title{\LARGE \bf
Optimal Multi-Manipulator Arm Placement for Maximal Dexterity during Robotics Surgery
}

\author{Mingwei Xu$^{1^*}$, James Di$^{2^*}$, Nikhil Das$^{1}$ and Michael C. Yip$^{1}$  \textit{Senior Member, IEEE}
\thanks{* Equal Contribution}
\thanks{$^{1}$Mingwei Xu, Nikhil Das, and Michael Yip are with Department of Electrical and Computer Engineering, University of California San Diego, La Jolla, CA 92093 USA
        {\tt\small \{m8xu, nrdas, yip\}@ucsd.edu}}%
\thanks{$^{2}$ James Di is with Department of Computer Science and Engineering, University of California San Diego, La Jolla, CA 92093 USA
        {\tt\small y2di@ucsd.edu}}
}

\begin{document}
\maketitle
\thispagestyle{empty}
\pagestyle{empty}

\begin{abstract}
Robot arm placements are oftentimes a limitation in surgical preoperative procedures, relying on trained staff to evaluate and decide on the optimal positions for the arms. Given new and different patient anatomies, it can be challenging to make an informed choice, leading to more frequently colliding arms or limited manipulator workspaces. In this paper, we develop a method to generate the optimal manipulator base positions for the multi-port da Vinci surgical system that minimizes self-collision and environment-collision, and maximizes the surgeon’s reachability inside the patient. Scoring functions are defined for each criterion so that they may be optimized over. Since for multi-manipulator setups, a large number of free parameters are available to adjust the base positioning of each arm, a challenge becomes how one can expediently assess possible setups. We thus also propose methods that perform fast queries of each measure with the use of a proxy collision-checker. We then develop an optimization method to determine the optimal position using the scoring functions. We evaluate the optimality of the base positions for the robot arms on canonical trajectories, and show that the solution yielded by the optimization program can satisfy each criterion. The metrics and optimization strategy are generalizable to other surgical robotic platforms so that patient-side manipulator positioning may be optimized and solved.
\end{abstract}

\section{Introduction}
Robotic laparoscopic surgery offers important advantages to the primary surgeon over manual laparoscopic surgery including improved ergonomics and dexterity. However, a challenge that must be contended with is the significant volume and weight of the robot, occupying a large portion of the sterile part of the operating room \cite{LIMITATIONS}. Key elements of successful design for surgical robotic systems include sufficient robot workspace and reachability for the surgeon inside the patient and adaptability to any operating room \cite{ROBOT-ASSISTED}. While the da Vinci surgical systems can satisfy these requirements, the positioning of the robot arms to optimize these key elements is left to the surgical staff during preoperative setup procedures, which requires trained surgical staff to remember and carefully follow certain guidelines. This preoperative time when the robot is being set up is one major limitation of robotic over non-robotic procedures \cite{LIMB}\cite{PLASTIC-SURGERY}\cite{colorectalresection}\cite{flapsurgery}. The considerations the surgical staff take during arm placement include minimizing collisions of the arms with other arms, the endoscopic camera, and other obstacles in the operating room while still allowing the primary surgeon a wide range of motion inside the patient. However, even with careful manual arm placement, instruments collisions during the procedure are still reported \cite{DOCKING}.

\begin{figure}[t!]
    \centering
    \includegraphics[trim={1cm 2cm 0cm 2cm}, clip,width=\linewidth]{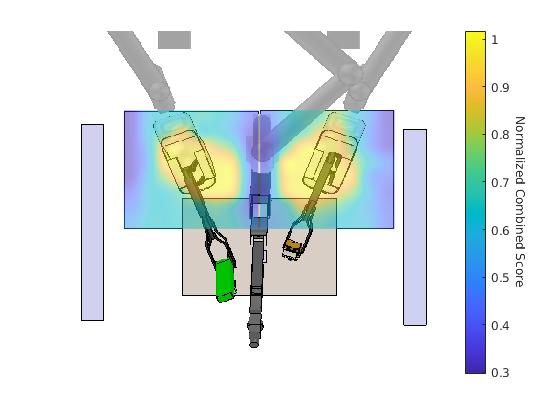}
    \setlength{\belowcaptionskip}{-17pt}
    \caption{Top-down view of environment and normalized scores of base positions for the da Vinci arms. The score for each $(x, y)$ position is the maximal score on the orientation around the $Z$ axis. The shown instrument arrangement is a simulated arrangement. }
    \label{top-down-view}
\end{figure}

Poor arm placement may cause them to collide with each other, or with equipments surrounding the patient \cite{DOCKING}. These collisions may cause issues during the operation including:

\begin{itemize}
    \item damage to the robot or surgical equipment \cite{6130913}, potentially increasing operation cost or time.
    \item compromising the sterility of the operating room (e.g., by moving separating curtains or disposable sterile arm drapes), increasing the chance of infection for the patient or surgical staff.
    \item physical limits of the surgical tools that minimize robot reachability in the body.
\end{itemize}

\begin{figure*}[t!]
    \centering
    \includegraphics[trim={0.5cm 5.25cm 0.5cm 3.00cm}, clip, width=1.0\linewidth]{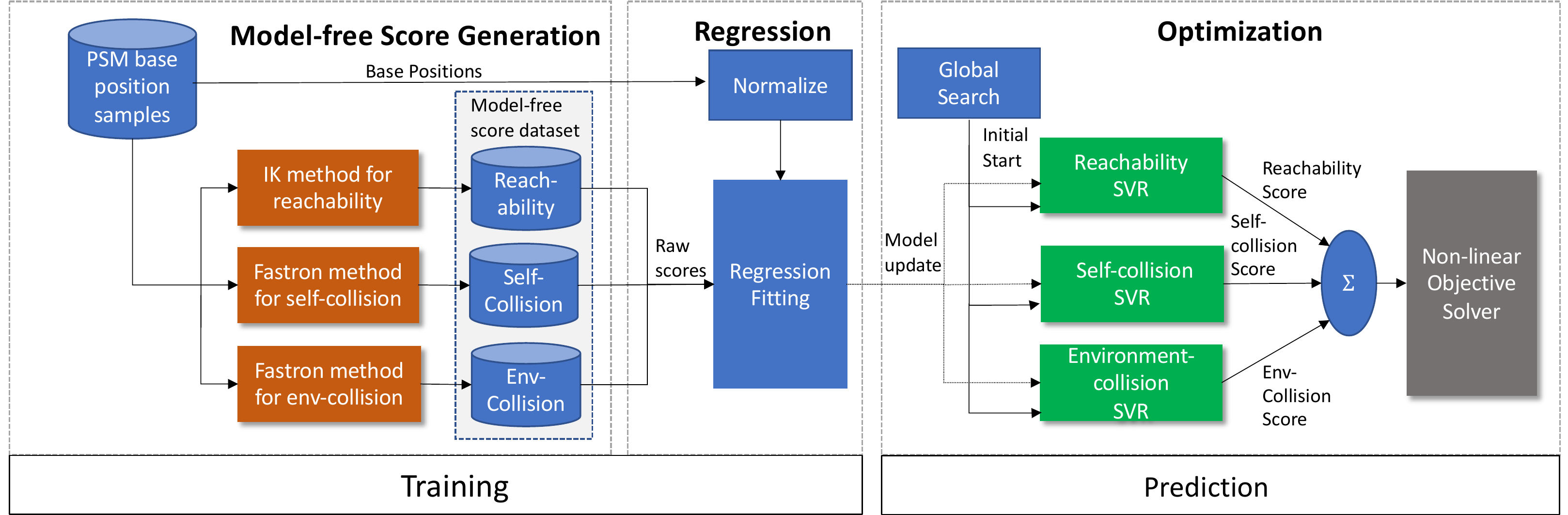}
    \setlength{\belowcaptionskip}{-10pt}
    \caption{\small{Data generation, regression and optimization pipeline. The samples of base position space $(X, Y, \Theta)$, and their three scores collected over IK and Fastron methods form the training dataset. Separate regressors are fitted over the normalized base positions to learn smooth and continuous score functions, with the three scores as targets. For prediction, global search with multiple starting candidates is used to find the optimal base position over a weighted sum of the three  scores.}}
    \label{fig:pipeline}
\end{figure*}

This paper aims to provide an assistive solution for optimal-positioning of patient-side manipulators (PSMs), and a method that analyzes and minimizes potential collision of the PSMs with themselves and other operating room obstacles, while maximizing the surgeon’s workspace manipulability. The output of the proposed method is a suggested set of PSMs base setup positions for the surgical robot staff, along with a score representing the optimality of the suggested solution. The optimal positioning is demonstrated on the multi-port da Vinci surgical system. An example of the score over the base positions is illustrated in Fig. \ref{top-down-view}. Note that analysis and experiments in this work are performed on the da Vinci Research Kit \cite{dvrk} and are not necessarily transferable to da Vinci X/Xi systems due to their geometry differences.


\subsection{Contribution}



The two main contributions of the paper are as follows:

\subsubsection{Definition of scoring functions to evaluate PSM placement metrics}
We define key metrics in PSM placement - self- and environment-collisions and intracorporeal manipulability, and hereby define scoring functions to analyze the degree of feasibility of different PSM setup positions. We then develop a pipeline to generate a numerical dataset that represents costs of various PSM setups, which can be utilized in machine learning and optimization methods.

\subsubsection{Development and evaluation of an optimization technique for finding optimal base position with pre-defined trajectories}
Using the scoring functions described above, we develop an optimization method for finding the base position, and demonstrate its effectiveness on pre-defined trajectories.

\subsection{Related Work}
Despite the lack of research towards optimal PSM arm placement, there are several works focusing on da Vinci system port placement, which is another critical step in the system setup with reachability and collision constraints. In addition, these works are primarily concerned with self-collisions but do not consider environmental obstacles.


Considering the optimal port placement problem, Stricko et al. \cite{stricko2018port} describe independent scores representing reachability, spar interference, and naturalness of the port placements. A geometric method is mainly used to perform score analysis, where they approximate the PSM workspace as cones and spar (the PSM link controlling the insertion axis) bodies as cylinders. However, they simply iterate through different port placements samples and find the maximum weight score, which can be limited by sample number and hard to reach the global optimal. Extracorporeal collisions are not considered in this work. In \cite{4399354}, Sun et al. try to solve the problem via optimization techniques. The authors define a procedure-specific optimization technique aiming for maximum reachability and visibility, and minimal collision inside the patient. Nevertheless, techniques to analyze and optimize for reduced environment-collisions are neglected.

Towards workspace analysis, Liu et al. \cite{6130913} define performance indices for instrument dexterity and manipulability based on singular values of the manipulator Jacobian. These indices are used to find optimal port placements but environment-collisions are not taken into consideration in optimization.


\section{Methods}
We describe a method to minimize PSM self-collisions both inside and outside of patient's body, minimize environment-collisions, and maximize the PSM reachability in the operative workspace inside patient's body. We consider the positions of the base of each PSM, where each position is represented with three elements: $(x, y, \theta)$, where $x$ and $y$ are translational positions and $\theta$ is the orientation about the vertical $Z$ axis. The $Z$ axis is excluded from the state space as the height of the base is constrained by the patient position/geometry.

We define three score functions that represent the self-collision, environment-collision, and reachability objectives. We then use a regression model to fit a smooth and continuous model to a set of these scores and ensure efficient calculation of these scores. Finally, optimization techniques are employed using a combination of these model-based scores as an objective function. The entire pipeline is shown in Fig. \ref{fig:pipeline}.

\subsection{Model-free Score Generation}

In the interest of optimizing PSM base setup positions with regard to self-collisions, environment-collisions, and reachability of the PSM instruments, we must define representative scores that characterize the degree to which these objectives are met for a given PSM base position. In this section, we define three score functions that are based on estimated proportions of instrument positions that are self-collision-free, environment-collision-free, or able to reach certain targets in the operative field. These estimated proportions are based on uniformly sampled instrument positions, but in practice, the distribution of these instrument positions may be data-driven to better reflect typical positions during real surgical procedures. The pipeline to acquire a set of these scores is summarized in the left-most block in Fig. \ref{fig:pipeline}.

\subsubsection{Reachability Analysis and model-free score Calculation}
Reachability is a fundamental metric when evaluating the optimal PSM base setup position. A good setup should maximize the PSM end-effectors’ reachability within the region of interest to enhance intracorporeal manipulability.

To analyze reachability, the target region is uniformly divided into voxels. For a given PSM base position, each voxel center is considered as a target position for the PSM end-effector. Here, given a PSM base position, its remote center of motion (RCM) is fixed. Iterating through each voxel target, we solve inverse kinematics (IK) for each PSM arm with respect to its yaw, pitch, and insertion joints, then check if the obtained solution is within joint limits. If so, the corresponding target region will be marked as reachable. In our experiment, we use the damped least-squares method \cite{4543501}, a numerical method with greater stability around singularities \cite{Buss2004IntroductionTI} to obtain IK solution. While numerical IK solvers are sometimes vulnerable to local optimums \cite{1688527}, this is not observed during our experiment due to the non-redundant system setup and the characteristic of PSM arm.

In this way, reachable voxels can be queried, which creates an intersecting set of target region space and PSM arm end-effector reachable workspace. The reachability model-free score is calculated by dividing the number of the queried target region voxels by the total number of the target region voxels, which physically approximates the volume portion of the target region where PSM arm end-effector is able to reach.

\begin{equation}
    score_{reach}(x, y, \theta) = \frac{\text{\# reachable targets}}{\text{\# targets}}
\end{equation}

Reachability score ranges from 0 to 1, where 0 means the PSM arm end-effector cannot reach within the target region at all, and 1 means the PSM arm end-effector reachable workspace contains the whole target region space. The score is analyzed individually for each PSM arm’s setup positions.

In practice, voxelization of the workspace and the cubic workspace assumption are not required. Instead, targets can be generated based on imagery from the endoscopic camera or preoperative patient scans, while the reachability score can be calculated using the same approach. Comparing to commonly-used geometric methods for reachability analysis, the sampling method is more flexible and adaptive in this data-driven approach in practice.

\subsubsection{Collision Analysis and Model-free Score Calculation}
In order to minimize the potential for PSM self- and environment-collisions, a collision detection algorithm must be implemented. Due to the significant number of collision checks to evaluate and the large computational cost of performing geometric collision checks, we use Fastron \cite{DBLP:journals/corr/abs-1709-02316}, as a collision classifier model which approximates geometric collision checks with a binary classifier learning from labeled collision and non-collision configurations. This method generally provides significant computational speed improvement to traditional collision checking. A geometric collision checker is still needed to train the Fastron model and generate datasets for Fastron when evaluating different PSM setups, and here we use the V-REP collision checker.

The Fastron algorithm creates a nonparametric model that serves as a faster proxy to geometric collision checking. The input into the Fastron model is a robot configuration $q$, and the output is its predicted collision status. The model $F(q)$ is a sparse sum of kernel functions $\{k(x_i, q)\}$ as shown below, with weights $\{\alpha_i\}$ determined through a variant of gradient descent,
\begin{equation}F(q)=\sum^{N}_{i=1} k\left(x_{i}, q\right) \alpha_{i}\end{equation}
where $\textit{sign}(F(q))$ denotes the collision status for a robot in configuration $q$.  We refer the readers to \cite{DBLP:journals/corr/abs-1709-02316} for the derivation of the Fastron model.

For self-collisions, each PSM arm is fitted with a collision classifier independently, and collisions are checked between PSMs the Endoscopic Camera Manipulator (ECM). In addition, another collision classifier for each PSM arm is fitted towards classifying environment-collisions.

Collision scores are evaluated at each PSM base position. When PSM bases are set up, we randomly sample 1,000 PSM arm joint configurations within joint space for each PSM arm, check self- and environment-collision for each configuration. The collision score at each position is calculated by dividing collision-free configurations by total configurations. In this way, we get the collision-free score ranging from 0 to 1, where a score of 0 represents the maximum in-collision risk and a score of 1 represents the minimum in-collision risk. The equations for self-collision scores and environment-collision scores are thus constructed as follows:

\begin{equation}
    score_{self}(x, y, \theta) =  \frac{\text{\# self-collision-free positions}}{\text{\# instrument positions}}
\end{equation}
\begin{equation}
    score_{env}(x, y, \theta)  =  \frac{\text{\# environment-collision-free positions}}{\text{\# instrument positions}}
\end{equation}

The model-free score datasets are hereby constructed, which contains each PSM base’s setup $(X, Y, \Theta)$, and the label as its reachability score, self- and environment-collision-free score separately.

\vspace{-1pt}
\subsection{Regression and Optimization}
In this section, a method is described that uses the scoring functions in Section A and finds a differentiable regressor that is then used for the base positions of the robot. We first sample sparsely in the $(X, Y, \Theta)$ space of the base position, and obtain the three scores for each position. Then the configurations along with their scores are used as the training dataset for fitting regression models to obtain continuous and smooth costmap for each criterion. These smooth costmaps are used in order to perform continuous gradient descent to find local optimal solutions efficiently, rather than a sample-only method that would be computationally and memory burdensome. We use Support Vector Regression Machines \cite{drucker1997support} as our regression model for the sparsity of the model, as well as its training efficiency compared with other regression models on our datasets.

For each score, we fit a SVR model to minimize the regression loss between the predicted score and the true score. Each dimension of our base positions is first normalized to $[-1, 1]$, and then used as the feature for regression. We use $y(x_i)$ as the predicted score for each normalized configuration $x_i$ given a scoring function. The score-prediction rule for each configuration $x$ can be written as
\begin{align}
     y(x) = \sum_{i = 1}^{N} (a_i - \hat{a_i})k(x, x_i) + b \label{eq:prediction}
\end{align}
where $a_i$ and $\hat{a_i}$ are the Lagrange multipliers. The Gaussian kernel $k(x, x_i) = exp(-\gamma\norm{x_i - x}^2)$ is used to map the features to a higher dimensional space. $\gamma$ is a pre-defined parameter controlling the width of the kernel. The learn-able parameters in Eq.\ref{eq:prediction} are the multipliers $a_i, \hat{a_i}$ for each datapoint $x_i$, as well as the bias $b$. Only the support points have non-zero $a_i$ or non-zero $\hat{a_i}$. We refer the readers to \cite[Ch.5]{bishop} for a complete derivation of the SVR model.

To find the optimal base position given three estimated score maps of the base positions, we use a global optimization technique to optimize the base positions with respect to the objective function. As each original model-free score is within $[0, 1]$ and the objective is minimizing the regressson loss, it is possible for the predicted scores to lie outside of this range. To ensure that each of the three scores is compared equally, we use a clipping function to clip the predicted score into $[0,1]$ and obtain $\textit{score}'(x)$ for each metric. We experimented with scaling the predicted scores using a smoother function, but found empirically that using the clipping function produces more highly-scored base positions when considering all three criteria. Weights may be applied to each of the scores to prioritize different objectives. The objective function can thus be formulated as: \\
\begin{equation}
\begin{aligned}
& \underset{x}{\text{maximize}}
& & f(x) = \text{w}_{\text{reach}} \cdot \text{score}_{\text{reach}}'(x) + \text{w}_\text{self} \cdot \text{score}_{\text{self}}'(x) \\
& & & + \text{w}_\text{env} \cdot \text{score}_{\text{env}}'(x)\\
& \text{subject to}
& &  x \in [-1, 1] ^ 3
\end{aligned}
\end{equation}
\indent
Since the weighted sum of the scores is non-linear due to the kernel mapping in SVR, it is highly likely for the scores to have a local minimum. To compensate for this we use a global optimizer that sample multiple starting candidates for base positions in the range of [-1,1] as initialization to the optimization problem, and later scale the solution back to its joint limits range.


\section{Simulator, Environment, and Setup}
\subsection{Simulator}
For our study a simulated environment is created based on the da Vinci Research Kit implemented in the V-REP simulator by Fontanelli et al. \cite{8487187}. V-REP has built-in kinematics solvers and collision detectors that can be utilized via its Application Programming Interface (API), which are used in our experiment. This part of the simulation model is generally interchangeable with other versions of the da Vinci systems, or other surgical robotic platforms for that matter.

\vspace{-3pt}
\subsection{Environment}
The simulated environment contains the da Vinci system with two PSM arms and one ECM arm, along with a representation of its setup joints. While a third PSM arm is available in some da Vinci systems, it is not used in all da Vinci multi-port procedures and oftentimes tucked away during the surgery, thus not included in this environment.

\paragraph{Environment Setup for Reachability Analysis}
For reachability analysis, the target region for the PSM end-effectors is described by a cube with 6 cm side length. In general, any arbitrary reachable volume may be chosen in practice, and may be defined based on an atlas of procedures and patient morphologies.

\paragraph{Environment Setup for Collision Analysis}
For collision analysis, PSM arms are meshed with convex hulls by components. The ECM mesh contains two parts, where the first part is a collection of convex hulls of ECM arm components, and the second part is a cone fit to the ECM endoscope that represents its movement workspace during operation. The ECM cone vertex is at the remote center of ECM, and aligned with the centerline of the endoscope. The cone shape is chosen to represent the ECM workspace because the ECM might re-orient in any direction during the operation. A fully retracted ECM is used to define the cone to estimate the largest volume of potential extracorporeal collisions. For self-collisions between arms, collisions are detected pairwise between each PSM mesh and the ECM mesh. These geometric descriptions of the arms account for both intracorporeal and extracorporeal collisions. 

Two planes on the left and right side of the da Vinci system construct a simple case for environment-collision, representing obstacles such as walls, other equipment carts, or curtains. Environment-collisions are detected pairwise between each PSM mesh and plane.

\paragraph{Improve Collision Detection Efficiency}
As self- and environment-collision scores require numerous collision checks, we strive to make the collision checking routine as efficient as possible. While a learning-based collision detector like Fastron is used to further improve collision check speed, it also requires a geometric collision checker to generate training data.

We compare to V-REP's built-in collision checker \cite{6696520} in our experiments, which has a remote API that provides simple usage for collision detection; however, it can only perform one collision check per API call, causing communication lag to dominate the time consumption of millions of collision checks during our experiment. To minimize communication lag, we wrote custom API functions embedded inside the V-REP simulated environment, which perform up to 1,000 collision checks per function call, limited by V-REP buffer size. This method is also used to generate training data for Fastron.

Using Fastron, we achieved even higher collision check speed, which is about 2 times faster than the V-REP collision detector with custom API, as shown in Table \ref{collision_check_timing}. Therefore scores for self- and environment-collisions may be computed faster.

As Fastron is an approximate method for collision checking, we must also evaluate its accuracy relative to V-REP’s geometric collision checking. In our experiments, Fastron achieves an accuracy of 98.5\%, a true positive rate of 99.4\%, and a true negative rate of 98.1\%. While the lack of 100\% accuracy can lead to potential overestimates (1-2\%) of the effective workspace of the robot, since the method examines optimal base positions as a whole across millions of samples, this 1-2\% is only important if competing base position solutions are very similar to one another, meaning that any of those base positions would be reasonable.

\begin{table}[h!]
\centering
\begin{tabular}{ c|c } 
\hline
\textbf{Method} & \textbf{Time per Collision Check (ms)} \\
\hline

V-REP (remote API) & 12.8 \\
V-REP (custom API) & 1.26 \\
Fastron & 0.455 \\
\hline
\end{tabular}
\setlength{\belowcaptionskip}{-20pt}
\caption{Self-collision detection timing for different collision detectors}
\label{collision_check_timing}
\end{table}


\section{Results}
\subsection{Model-Free Score Results}
Reachability score performance is evaluated in the simulated environment, where the PSM tools are expected to reach inside the defined target region. When generating the raw model-free scores, the PSM arm base position is fixed on the Z-axis, where the joint center connecting the PSM arm base and setup joint base is at 0.66 m height. Then, two PSM arm bases are traversed through two 0.7x0.7 m search grids on the XY plane, where the grid centers are the default base setup in the the simulated da Vinci system set by Fontanelli et al \cite{8487187}. The orientation $\theta$ is constrained to be between $\pm 0.3$ rad.

For self- and environment-collision-free scores, we tested the V-REP collision detector using our custom API and the Fastron collision detector. Self-collision is checked between PSM arm meshes and ECM arm meshes including the endoscope workspace. Environment-collision is checked between PSM meshes and planes on the left and right sides of the da Vinci system.

As Fastron is an approximate collision checker, MSE is evaluated using the geometric collision checker in V-REP as ground-truth. For a set of configurations $\{q_i\}$, $\text{MSE}(s, \hat{s}) = \frac{1}{n}\sum_{i}^{n} (s_i - \hat{s_i})^2$, where $\{s_i\}$ are the true scores and  $\{ \hat{s_i}\}$ are the estimated scores. The result in Table \ref{fastron_timing} indicates that using Fastron as the collision detector can highly improve raw collision score calculation speed while giving a very similar score results as when using the V-REP collision detector.

For all model-free score calculations, the timing result represents the average score calculation time for each two PSM arm base setups $(x_{PSM1}, y_{PSM1}, \theta_{PSM1}, x_{PSM2}, y_{PSM2}, \theta_{PSM2})$, and it is obtained by averaging the total calculation time of 700 different two PSM arm base setups.


\begin{table}[t]
    \vspace{5pt}
    \centering
    \begin{adjustbox}{width=0.95\linewidth}
    \begin{tabular}{ l l l l }
\hline
                    & \textbf{Method} & \textbf{Timing per Setup Evaluation (s)} & \textbf{MSE} \\ \hline
$score_{reach}$       & IK & 1.42 & NA \\ \hline
$score_{self}$ & V-REP (custom API) & 1.29 & NA \\
                    & Fastron & 0.501 & 0.003 \\ \hline
$score_{env}$ & V-REP (custom API) & 0.300 & NA \\ 
                    & Fastron & 0.151 & 0.003 \\ \hline
\end{tabular}
    \end{adjustbox}
    \setlength{\belowcaptionskip}{-15pt}
    \caption{Timing and Mean-Squared Error(MSE) for model-free score calculation.}
    \label{fastron_timing}
\end{table}

\begin{figure}[h!]
    \vspace{5pt}
    \centering
    \begin{subfigure}{.47\linewidth}
        \includegraphics[width=\linewidth]{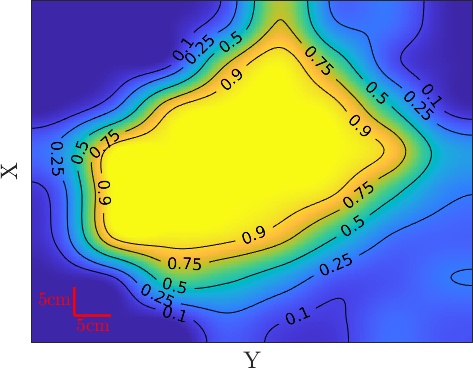}
        \caption{Reachability}
        \label{reachability}
    \end{subfigure}  
    \begin{subfigure}{.47\linewidth}
        \includegraphics[width=\linewidth]{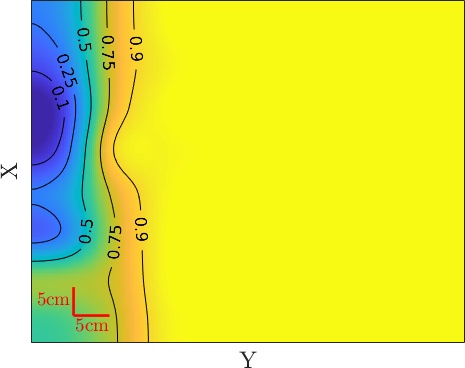}
        \caption{Self-Collision-Free}
        \label{self-collision-free}
    \end{subfigure}  
    \begin{subfigure}{.47\linewidth}
        \includegraphics[width=\linewidth]{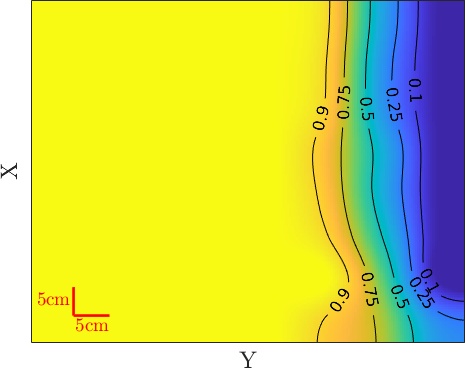}
        \caption{Env-Collision-Free}
        \label{env-collision-free}
    \end{subfigure}
    \begin{subfigure}{.47\linewidth}
        \includegraphics[width=\linewidth]{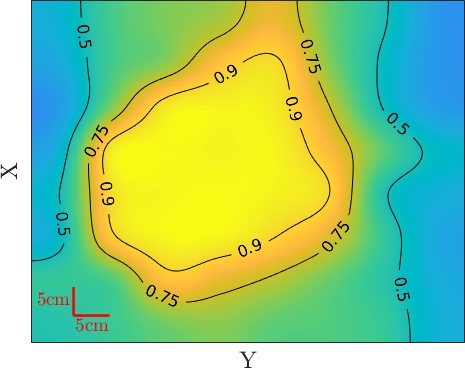}
        \caption{Combined}
        \label{combined}
    \end{subfigure}  
\setlength{\belowcaptionskip}{-10pt}
\caption{\small{PSM1 score results. The warmer the color, the higher the score for a base position $(x, y)$. For readability, the plots shown are for a specific value $\theta$.}}
\label{model-free score results}
\end{figure}

Example scores of a single PSM arm are plotted in Fig \ref{model-free score results}. As there are three explanatory variables $(X, Y, \Theta)$ and one response variable, the heatmap plot shows the maximum score along the $\Theta$ axis for each $X-Y$ pair from each PSM base setup, where the warmer the heatmap color, the higher score it can achieve among all possible $\theta$ angles at that $X-Y$ pair. Note how the reachability score is high close to the center and lower as the arms move away, how the proportion of self-collision-free configurations decreases closer to the center, and how the proportion of environment-collision-free configurations decreases closer to the walls in our environment.

\subsection{Trajectory Evaluation}

To evaluate the base position found via optimization in a realistic scenario, we define representative circular trajectories in the Region of Interest (RoI), and evaluate the aforementioned three metrics on the instrument configuration for each waypoint of the trajectory: reachability, self- and environment-collision-free. The canonical trajectories are representations of paths that surgeons would execute in the RoI. We average the metrics over the trajectories and obtain quantitative measures of the base positions.

\begin{figure}[h!]
     \centering
     \includegraphics[width=0.88 \linewidth]{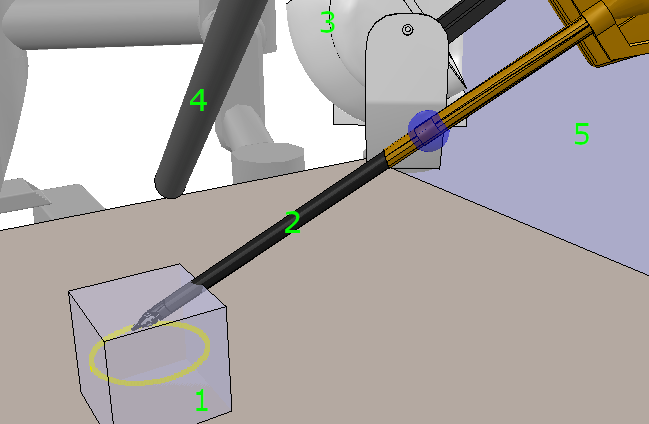}
     \setlength{\belowcaptionskip}{-10pt}
     \caption{\small{Exemplary trajectory for evaluating base position. 1.RoI, the yellow circle represents the trajectory the PSM end-effector tries to execute; 2.PSM Instrument; 3.PSM arm base; 4. ECM; 5. Obstacles.}}
     \label{trajectory for evaluation}
\end{figure}

For a single trajectory, we define $n$ waypoints on the trajectory, and use inverse kinematics to solve for the instrument configurations with the base position fixed. We check if the instrument satisfies the three criteria. To account for the different paths surgeons might execute, we include trajectories with 7 different locations and 3 axes of rotations, in total 9 different trajectories. An illustration of the trajectory being evaluated is included in Fig \ref{trajectory for evaluation}.

For reachability, a target waypoint of the trajectory is reachable if (a) the IK-solved instrument position is close to the target position within a threshold (less than half of 6mm); (b) the instrument position is within the RoI; (c) The solved instrument configurations satisfy the instrument joint limits. If the target position is reachable, we obtain the extra-corporeal collision statuses of the IK-solved instrument position. Otherwise, we compute the collision statuses of the instrument configuration that gives the closest end-effector position to the target position, as we envision that surgeons always try to execute the trajectory as closely as possible.

\vspace{-5pt}
\subsection{Base Position Evaluation Results}
\begin{figure}[h!]
    \vspace{5pt}
    \begin{subfigure}{.22\textwidth}
        \includegraphics[trim={1cm 1cm 1cm 8cm}, clip,width=\linewidth]{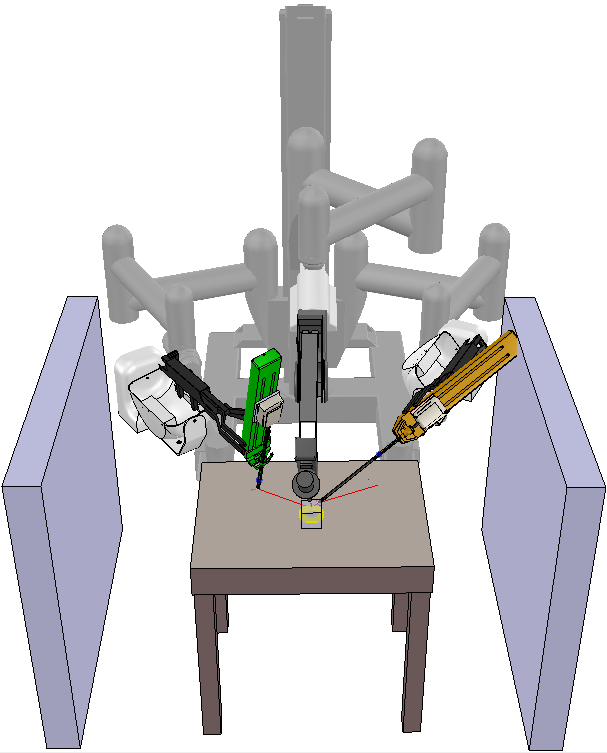}
        \caption{High self-collision-free, low reachability, environment-collision-free}
        \label{base_position_A}
    \end{subfigure}
    \hspace{8pt}
    \begin{subfigure}{.22\textwidth}
        \includegraphics[trim={1cm 1cm 1cm 8cm}, 
        clip, width=\linewidth]{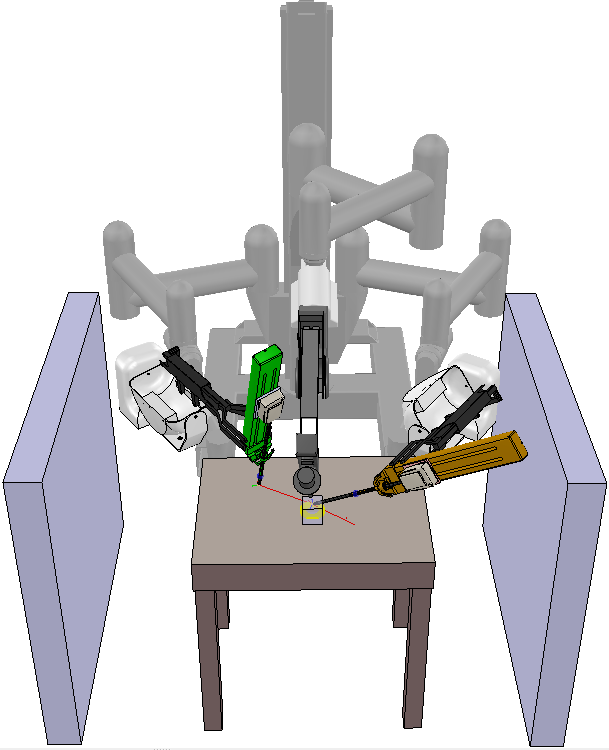}
        \caption{High reachability, environment collision-free, self-collision free}
        \label{base_position_B}
    \end{subfigure} 
    \setlength{\belowcaptionskip}{-15pt}
    \caption{\small{Suboptimal \& optimal base positions. For illustration, we show the hand-picked positions A from Table \ref{base_position_table} and one of the best optimized solutions B.}}
\label{base_position_figures}
\end{figure}

Table \ref{base_position_table} shows the trajectory scores with one hand-picked suboptimal base position (A), and four base positions with our optimization technique (B-E). A is sub-optimal because the reachability and environment-collision-free metrics are low when evaluated over the trajectories.Visualization of the corresponding base position can be found in Fig.\ref{base_position_A}.

For base positions B-E, we apply different weights on the three criteria, and use the optimization technique described in Method Section (B) to obtain the corresponding positions. We use fmincon as our local solver, which is a non-linear constraint optimization problem solver provided by MATLAB Statistics and Optimization toolbox \cite{MatlabOTB}. We evaluate our results with two global gradient-based solvers GlobalSearch and MultiStart. The results we report here use GlobalSearch solver, with 100 different initialization to find the global optimum. Notice that when the criteria have equal weights, we can obtain a solution that satisfies all three criteria for all waypoints in our trajectories (Table \ref{base_position_table} B). In addition, we observe that prioritizing reachability and environment-collision-free in our environment also lead to optimal positions (Table \ref{base_position_table} C\&E). Prioritizing self-collision-free (Table \ref{base_position_table} D) however, leads to a poor base position setup, as it penalizes base positions that are close to the RoI, which usually give a good reachability score and environment-collision-free score.

\begin{table}[h!]
     \centering
     \begin{adjustbox}{width=0.98\linewidth}
     \begin{tabular}{ c c c c c c c c}
\hline 

\textbf{Score} &  \textbf{A} & \textbf{B} &\textbf{C} &\textbf{D} & \textbf{E} \\\hline
$\alpha_{reach}$  &  N/A & $1.0$ & $5.0$ & $1.0$ & $1.0$ \\\hline
$\alpha_{self-collision}$ & N/A & $1.0$ & $1.0$  & $5.0$ & $1.0$ \\\hline
$\alpha_{env-collision}$ & N/A & $1.0$ & $1.0$ & $1.0$ & $5.0$ \\\hline
Reachability   &$0.00\pm0.00$ & $1.00 \pm 0.00$ & $1.00 \pm 0.00$ & $0.00 \pm 0.00$ & $1.00 \pm 0.00$\\\hline
Self-Collision-Free &$1.00\pm0.00$ & $1.00 \pm 0.00$ & $1.00 \pm 0.00$ & $1.00 \pm 0.00$ & $1.00 \pm 0.00$ \\\hline
Environment-Collision-Free &$0.29 \pm0.21$ & $1.00 \pm 0.00$ & $1.00 \pm 0.00$ & $1.00 \pm 0.00$ & $1.00 \pm 0.00$ \\\hline
Combined & $1.29\pm0.21$ & $3.00 \pm 0.00$ & $3.00 \pm 0.00$ & $2.00 \pm 0.00$ & $3.00 \pm 0.00$ \\\hline

\end{tabular}

     \end{adjustbox}
     \setlength{\belowcaptionskip}{-5pt}
     \caption{\small{Metrics evaluated over large workspace trajectories for hand-picked (A) and optimized (B-E) base positions.}}
     \label{base_position_table}
\end{table}

\section{Conclusion}
In this paper, a method is developed to determine optimal positions of the base of each PSM that minimizes self- and environment-collisions, and maximizes the surgeon’s reachability inside the patient. Scoring functions are defined over the $(X, Y,\Theta)$ dimensions of each PSM arm. We optimize extensively for fast query time for each metric by writing our own custom collision detection and IK scripts. Finally, a model-based optimization method is developed that uses the scoring functions and generates optimal base positions based on the weights of the criteria.

We collected empirical timing results of our APIs and demonstrate a 28 times speedup over the V-REP remote API by using Fastron, a collision-proxy. We evaluated our method extensively over pre-defined canonical trajectories in a RoI represented by a cube, and show that our optimization method could give an optimal position with respect to all three criteria. 

Future work will involve user studies on the physical da Vinci system for evaluating the suggested base positions, and incorporating real workspace geometries, extending beyond our cubic representation of workspace and planar external barriers.

\section{Acknowledgement}
This research was supported by the Intuitive Surgical Technology grant. We also thank Simon DiMaio, Omid Mohareri, Dale Bergman, and Anton Deguet for their assistance and support of the da Vinci Research Kit.





\bibliographystyle{ieeetr}
\bibliography{ms}

\end{document}